\documentclass[10pt,twocolumn,letterpaper]{article}

\usepackage{wacv}
\usepackage{times}
\usepackage{epsfig}
\usepackage{graphicx}
\usepackage{amsmath}
\usepackage{amssymb}
\usepackage{makecell}
\usepackage{hhline}
\usepackage{booktabs}
\usepackage{multirow}

\usepackage{adjustbox}
\usepackage{caption}
\usepackage[accsupp]{axessibility}
\usepackage{tabularx, booktabs}

\newcolumntype{?}{!{\vrule width 1.2pt}}
\newcolumntype{Y}{>{\centering\arraybackslash}X}

\def\wacvPaperID{720} 

\wacvapplicationstrack 

\wacvfinalcopy 

\ifwacvfinal
\usepackage[breaklinks=true,bookmarks=false]{hyperref}
\else
\usepackage[pagebackref=true,breaklinks=true,colorlinks,bookmarks=false]{hyperref}
\fi

\begin{document}

\title{Towards MOOCs for Lipreading: Using Synthetic Talking Heads to Train Humans in Lipreading at Scale}

\author{Aditya Agarwal$^*$\\
IIIT Hyderabad\\
{\tt\small aditya.ag@research.iiit.ac.in}
\and
Bipasha Sen$^*$\\
IIIT Hyderabad\\
{\tt\small bipasha.sen@research.iiit.ac.in}

\and
Rudrabha Mukhopadhyay\\
IIIT Hydearbad\\
{\tt\small radrabha.m@research.iiit.ac.in}

\and
Vinay Namboodiri\\
University of Bath\\
{\tt\small vpn@bath.ac.uk}

\and
C V Jawahar\\
IIIT Hyderabad\\
{\tt\small jawahar@iiit.ac.in}

}

\twocolumn[{
\maketitle
\begin{center}
\includegraphics[width=\textwidth]{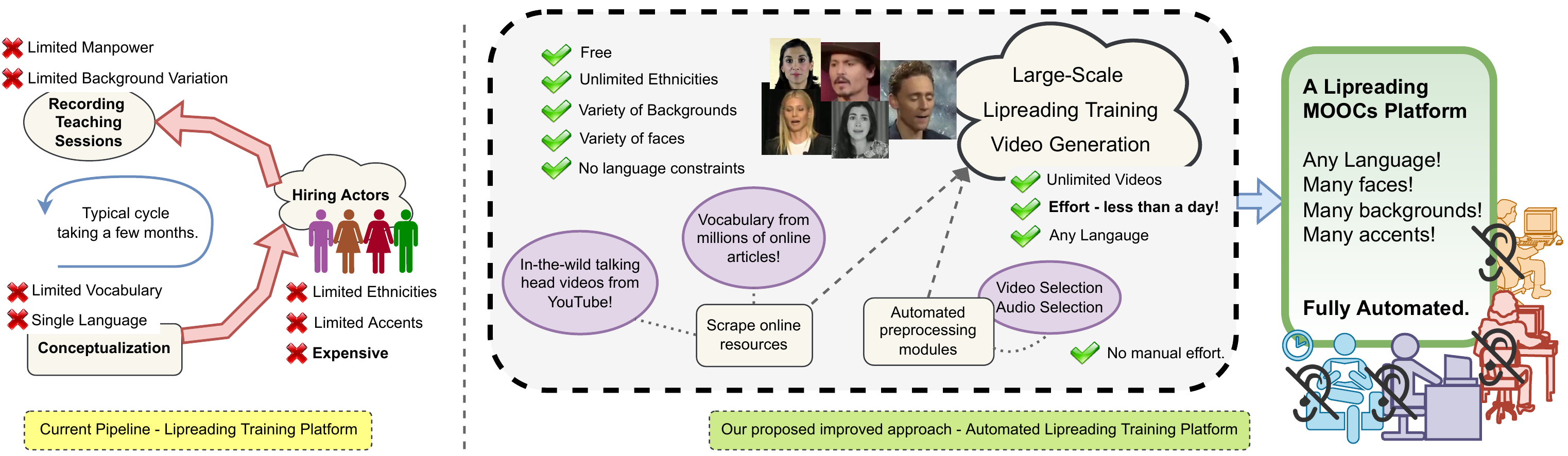}
\captionof{figure}{\small Lipreading is a primary mode of communication for people with hearing loss. The United States of America alone is home to 48 million people with some form of hearing loss. Despite these staggering stats, online lipreading training resources are scarce and available for only a handful of languages. However, hosting new lipreading training platforms is an extensive ordeal that can take months of manual effort. We propose a fully-automated approach to building large-scale lipreading training platforms. Our approach enables any language, any accent, and unlimited vocabulary on any identity! We envision a lipreading MOOCs platform to enable millions of people with hearing loss across the globe. In this work, we thoroughly analyze the viability of such an approach.}

\end{center}
}]
\def\thefootnote{*}\footnotetext{Equal contribution}
\thispagestyle{empty}

\begin{abstract}

Many people with some form of hearing loss consider lipreading as their primary mode of day-to-day communication. However, finding resources to learn or improve one's lipreading skills can be challenging. This is further exacerbated in the COVID$19$ pandemic due to restrictions on direct interactions with peers and speech therapists. Today, online MOOCs platforms like Coursera and Udemy have become the most effective form of training for many types of skill development. However, online lipreading resources are scarce as creating such resources is an extensive process needing months of manual effort to record hired actors. Because of the manual pipeline, such platforms are also limited in vocabulary, supported languages, accents, and speakers and have a high usage cost. In this work, we investigate the possibility of replacing real human talking videos with synthetically generated videos. Synthetic data can easily incorporate larger vocabularies, variations in accent, and even local languages and many speakers. We propose an end-to-end automated pipeline to develop such a platform using state-of-the-art talking head video generator networks, text-to-speech models, and computer vision techniques. We then perform an extensive human evaluation using carefully thought out lipreading exercises to validate the quality of our designed platform against the existing lipreading platforms. Our studies concretely point toward the potential of our approach in developing a large-scale lipreading MOOC platform that can impact millions of people with hearing loss. 
\end{abstract}


\section{Introduction}


Communication is a crucial ingredient that makes Humans the most intelligent species on the planet. While other animals also have different forms of communication, human language is more advanced by several orders of magnitude. But we are not inherently born with these skills! Then, how do we acquire them? Most of us learn linguistic skills through a formal education system consisting of schools, universities, and other organizations related to education. While this is still the most trusted \& popular way of imparting education, the 21st century has seen an exponential rise in online forms of education like the Massive Open Online Courses (MOOCs). Online courses are generally designed to cover hundreds of topics in various domains, including language, and are often available free of cost. MOOCs have several advantages over the physical form of education. They are more accessible, cheap, and reachable to a broader audience. In today's world, it is natural to learn a whole new language from the comfort of your home by attending a high-quality MOOCs course. 

Unfortunately, every person does not get the chance to learn linguistic skills like we usually do. Hearing loss is a common form of disability that can become a massive barrier to education! According to organizations like  WHO\def\thefootnote{$^1$}\footnote{\href{https://www.who.int/news-room/fact-sheets/detail/deafness-and-hearing-loss}{Deafness and Hearing Loss $|$ WHO}} and Washington Post\def\thefootnote{$^2$}\footnote{\href{https://www.washingtonpost.com/lifestyle/magazine/as-wearing-masks-becomes-the-norm-lip-readers-are-left-out-of-the-conversation/2020/05/22/23dcc4c0-8e19-11ea-a9c0-73b93422d691_story.html}{As wearing masks becomes the norm, lip readers are left out!}}, over 5\% of the world's population (432 million adults and 34 million children) and at least 48 million Americans are deaf with some form of hearing loss. About $500,000$ Americans have a disabling hearing loss that noticeably disrupts communication. 

\begin{figure}[t]
    \centering
    \includegraphics[width=0.83\linewidth]{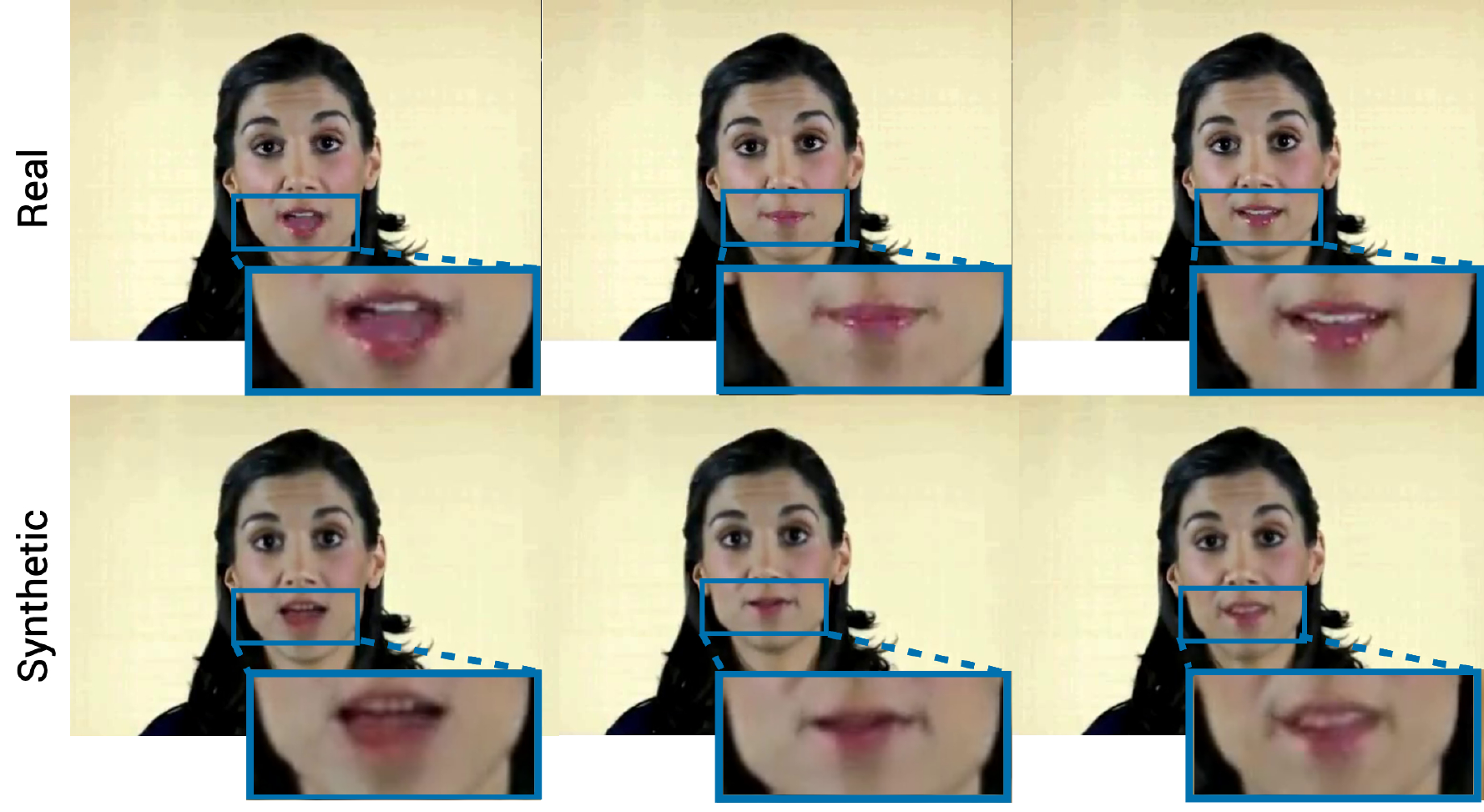}
    \caption{\small Talking-face video generated using our pipeline.}
    \label{fig:variations}
\end{figure}

Lipreading is a primary mode of communication for people with hearing loss. The Scottish Sensory Censor (SSC)\def\thefootnote{$^3$}\footnote{\href{http://www.ssc.education.ed.ac.uk/courses/deaf/ddec05f.html}{Factors which help or hinder lipreading $|$ SSC}} quotes, ``whatever the type or level of hearing loss, a child is going to need to lipread some of the time." However, learning to lipread is not an easy task! Lipreading can be thought of being analogous to ``learning a new language" for people without hearing disabilities. People needing this skill undergo formal education in special schools and involve medically trained speech therapists. Other resources like daily interactions also help understand and decipher language solely from lip movements. However, these resources are highly constrained and inadequate for many patients suffering from hearing disabilities. 

Inspired by the boom in online courses available for virtually every topic, we envision a MOOCs platform for \textbf{L}ip\textbf{R}eading \textbf{T}raining (LRT) for the hearing disabled. 

\paragraph{Current Online Platforms for LipReading Training} Platforms like lipreading.org\def\thefootnote{$^4$}\footnote{\href{https://www.lipreading.org/}{lipreading.org}} and lipreadingpractice\def\thefootnote{$^5$}\footnote{\href{https://lipreadingpractice.co.uk/Lip-Reading-Information/}{lipreadingpractice.co.uk}} provide basic online resources to improve lipreading skills. These platforms allow users to learn limited levels of lipreading constrained by resources. Unfortunately, the amount of vocabulary systematically covered during the exercises is extremely narrow. The videos also have minimal real-world variations in head-pose, camera angle, and distance to a speaker, making it difficult for a lipreader to adapt to the real world. Finally, since these resources are all available only in American or British-accented English, it becomes challenging for people from other regions to adapt to their local accents and languages. All the above factors severely limit the quality of human training. Therefore, we believe it is quintessential to scale the current lipreading training platforms to incorporate extensive vocabulary and introduce variation in videos, languages, and accents. However, recording videos is a costly affair. It requires expensive camera equipment, studio environments, professional editors, and a substantial manual effort from the perspective of a speaker whose videos are being recorded.

\begin{figure*}[t]
\centering
  \includegraphics[width=0.93\textwidth]{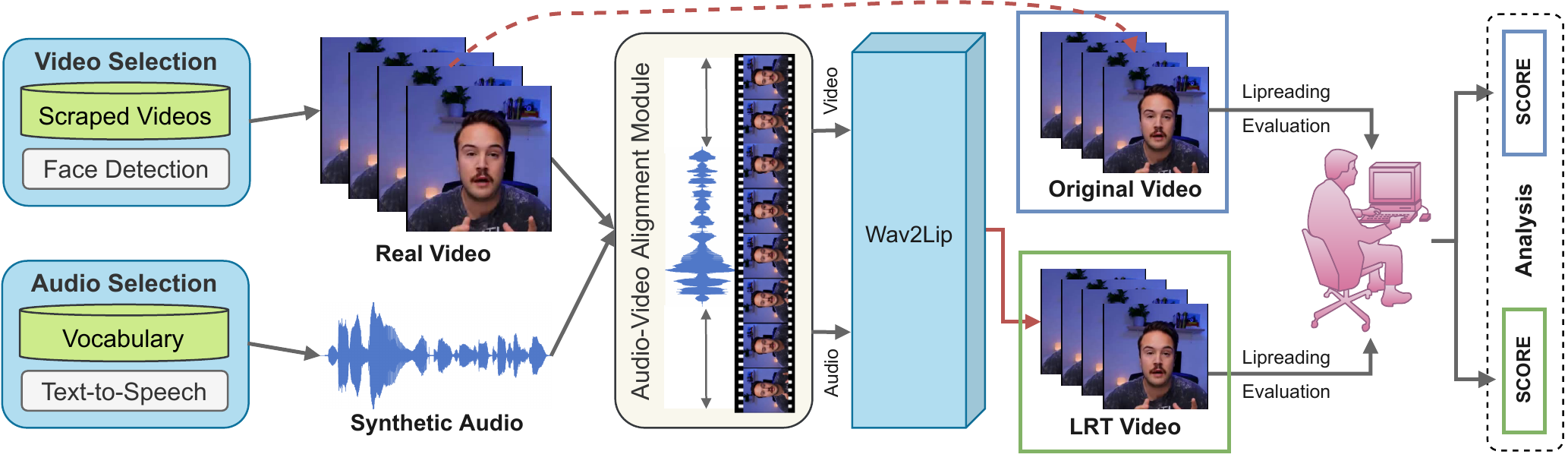}
  \caption{\small Proposed pipeline for generating large-scale lipreading training platform: \textbf{(a) Video Selection}: Videos are scraped from various online sources (such as YouTube), and invalid videos are filtered out. \textbf{(b) Audio Selection:} Synthetic speech utterances are generated using vocabulary curated from various online articles. \textbf{(c) Audio-Visual Alignment Module}: A video and a speech utterance is selected and aligned on each other such that the speech utterance overlaps with the region in the video with lip movements.
  \textbf{(d) Wav2Lip: } A state-of-the-art talking head generation model that modifies the lip movements of the video according to the speech utterance. \textbf{(e) User Evaluation: } A validation step to ensure that users perform comparably on real videos and synthetic videos generated using our approach.}
  \label{fig:main_pipeline}
\end{figure*}

To resolve this issue, we approach this from a different angle and ask: ``Can we replace real talking head videos used for training people suffering from hearing loss with synthetic versions of the same?" A synthetic talking head with accurate lip synchronization to a given text or speech signal can enable the scaling of LRT platforms to more identities, accents, languages, speed of speech, etc., making the training process more rigorous. We take advantage of the massive progress made by the computer vision community on synthetic talking head generation and employ a state-of-the-art (SOTA) algorithm~\cite{wav2lip}, as mentioned below.  

We propose a novel approach to automatically generate a large-scale database for developing an LRT MOOCs platform. We use SOTA text-to-speech (TTS) models~\cite{fastspeech2} and talking head generators like Wav2Lip~\cite{wav2lip} to generate training examples automatically. 
Wav2Lip~\cite{wav2lip} requires driving face videos and driving speech segments (generated from the TTS in our case) to generate lip-synced talking head videos according to the driving speech. It preserves the head pose, background, identity, and distance of the person from the camera while modifying only the lip movements, as shown in Fig.~\ref{fig:variations}. 

Our approach can exponentially increase the amount of online content on the LRT platforms in an automated and cost-effective manner. It can also seamlessly increase the vocabulary and the number of speakers in the database. We investigate the implications of our system for a range of deaf users and perform multiple experiments to show its effectiveness in replacing the manually recorded LRT videos. 
We show through statistical analysis that (1) the users' performance on lipreading videos is not significantly different when switching from `real' to `generated' videos, and (2) the benefit of lipreading platforms in one's native accent through an extensive user study. We believe our approach to generating fully synthetic videos is the first step towards developing an LRT MOOC platform to benefit millions of users with hearing loss.

\section{Related Work}
The usefulness of MOOCs as a medium of education has been accepted~\cite{harvard_business_review_2020} worldwide. Surveys like~\cite{moocs_1} analyze various aspects of the impact of MOOCs and help us understand their positives and negatives. MOOCs are shown to increase the audience and offer viable alternatives to the traditional form of education in~\cite{moocs_survey}. The increasing demand for content has also improved student engagement~\cite{moocs_2, moocs_wacv}. The requirement for MOOCs and other forms of online education has skyrocketed since the beginning of the COVID19 pandemic\def\thefootnote{$^6$}\footnote{\href{https://www.weforum.org/agenda/2020/04/coronavirus-education-global-covid19-online-digital-learning/}{The rise of online learning during the COVID-19 pandemic}}. We believe this trend to continue and impact different types of education required by people with special needs. Our work also aligns with assistive technology, where Digital media has historically played an important role. Much of these efforts have been invested in improving the communication skills of certain groups. In 2006, ~\cite{massarobosseler2006} published their work on ``Baldi", a computer-animated tutor to teach children with autism. Following this, another work~\cite{3dpronunciationtutor} has focused on generating 3D animated tutors for autism-affected children to improve their communication skills. Research aimed at improving the communication skills of the hearing impaired is also popular. \cite{barker2003} developed a computer-assisted vocabulary for educating the deaf to communicate orally. Special courses~\cite{hcihearing3deafstudycourse} are designed to help people with limited hearing abilities. Human-computer interaction interfaces~\cite{hcihearing1, virtual-avatar} targeted for similar groups are also prevalent. Recently a landmark work~\cite{sign_language_asst_2} targeted to create a home assistant for hard-of-hearing people. Their work mainly focused on incorporating sign language-based commands into a personal assistant. Similar efforts were made for automatic lipreading in~\cite{bmvc_lipreading, Ma_2021_WACV}.

\begin{figure*}
    \centering
    \includegraphics[width=0.85\linewidth]{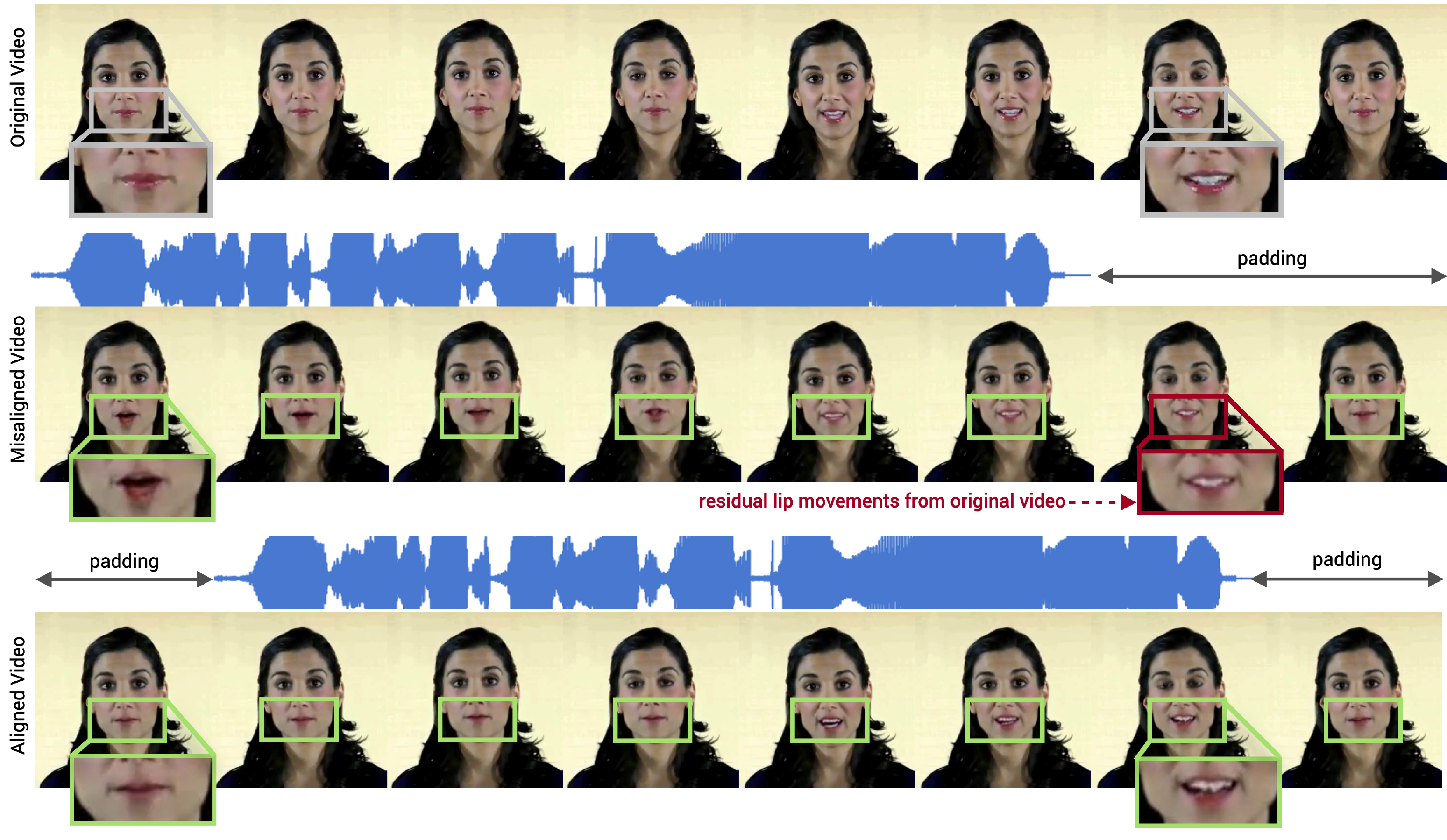}
    \caption{\small Audio-Video Alignment Module:
    Lip-sync models such as Wav2Lip modify the lip movements of an `Original Video'  (driving video) according to a given speech utterance. However, naively aligning the audio and video before passing through Wav2Lip can result in a `Misaligned Video' with residual lip movements, as indicated in red boxes. We design an audio-video alignment module that detects the mouth movements in the original video. We then align the speech utterance on the region with the mouth movements and add silence around the aligned utterance. Wav2Lip then generates an `Aligned Video' without any residual lip movements, as indicated in green boxes.}
    \label{fig:aud-vid-algn}
\end{figure*}

\section{Synthetic Talking Head Database}
\label{sec:curating_database}
Our lipreading training database generation pipeline: (1) Scrapes a set of face videos automatically from the internet. This helps us cover a large number of identities, background variations, lip shapes, etc. (2) Post-processes the scraped videos to filter out invalid faces (such as drastic pose changes). (3) Automatically curates a vocabulary of many words and sentences from various online sources. (4) Generates synthetic speech utterances on the curated vocabulary. (5) Selects a driving face video and a speech utterance to generate synthetic talking head videos using a SOTA talking head generation model, Wav2Lip, in our case. Wav2Lip modifies the lip movements of the driving video according to the speech utterance. The rest of the video (background, pose, etc.) is retained. These synthetic videos (with or without speech) are used to train humans in lipreading. The overall pipeline is illustrated in Fig.~\ref{fig:main_pipeline}.

\paragraph{Text-to-Speech System}
\label{subsubsec:tts}
We evaluate several TTS models: 
Fastspeech2~\cite{fastspeech2}, Real time voice cloning~\cite{rtvc},  Glow-tts~\cite{glowtts}, and Tacotron2~\cite{tacotron2} trained on LibriTTS~\cite{libritts} and LJSpeech~\cite{ljspeech17}. We evaluate them at  different speeds - 1$\times$, 1.5$\times$, 1.7$\times$, 2$\times$, pitch, and volume variations. We collect qualitative feedback from $30$ participants without any hearing loss on the clarity of the generated speech and report the Mean Opinion Scores (MOS) in the supplementary. 
For our experiments conducted on American-accented English, we use Fastspeech2 with 1$\times$ speed configuration pretrained on LJSpeech. For Indianised English accent, we use an online TTS\def\thefootnote{$^7$}\footnote{\href{http://ivr.indiantts.co.in/en/}{http://ivr.indiantts.co.in/en/}} with qualitatively similar performance to the speech generated by FastSpeech2. The TTS models used in our pipeline are configurable plug-and-play modules and can be replaced with any other TTS. This allows scalability and variations with little to no manual effort.

\begin{figure*}[t]
    \centering
    \includegraphics[width=0.83\textwidth]{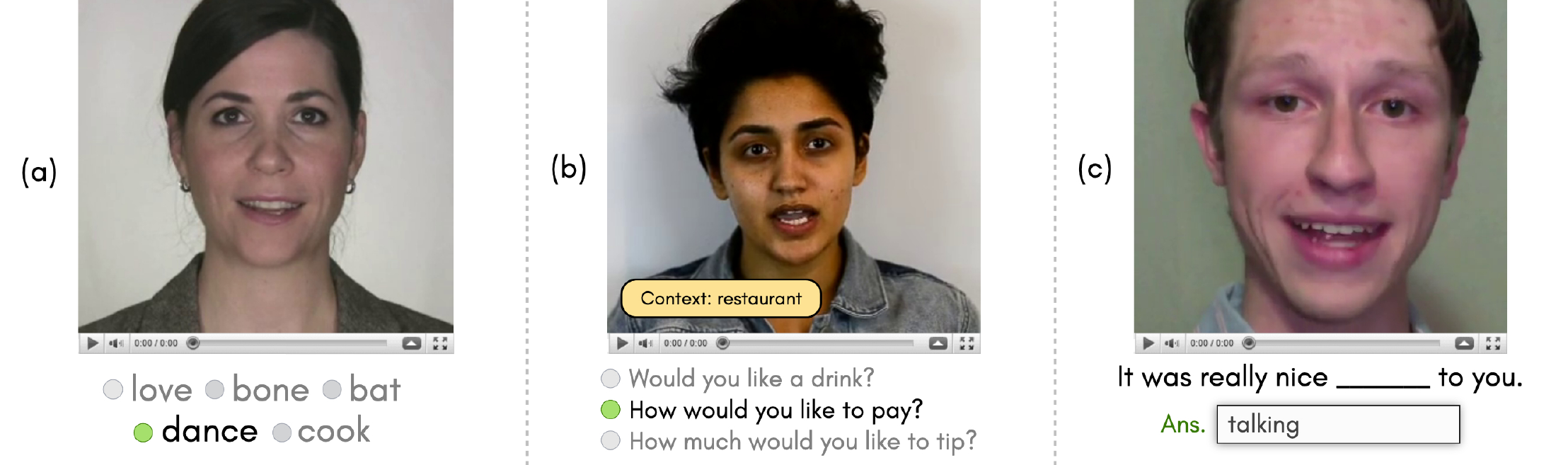}
    \caption{\small Examples of different protocols used for our user study. (a) lipreading isolated words (WL): the speaker mouths a single word, and the user is expected to select one of the multiple choices presented. (b) lipreading sentences with context (SL): the speaker mouths an entire sentence. The user is presented with the context of the sentence and is expected to select one of the sentences in multiple choices, and (c) lipreading missing words in a sentence (MWIS): the speaker mouths an entire sentence. The user is presented with a sentence with blanks (masked words); the user needs to identify the masked word from the video and sentence context and answer in text format.} 
    \label{fig:quiz_examples}
\end{figure*}

\paragraph{Synthetic Talking Head Videos}
\label{subsec:wav2lip}
Since 2015, talking head generation models that modify the lip movements according to a given speech utterance have gained much traction in the computer vision community~\cite{obamanet, text-based-editing, Yao2020IterativeTE}.
While some of these works generate accurate lip-sync, they are trained for specific speakers requiring large amounts of speaker-specific data. \cite{3d_tf} can be remodeled for generating talking heads but require far more manual intervention limiting their use in our approach. Recent advances like LipGAN~\cite{lipgan} and Wav2Lip~\cite{wav2lip} are perfect for our approach since they work for any identity without requiring speaker-specific data.
Consequently, we adopt Wav2Lip in our pipeline. Wav2Lip takes a face video of any identity (driving face video) and audio (guiding speech) as inputs. The model then modifies the lip movements in the original video to match the guiding speech, as shown in Fig.~\ref{fig:variations}. The rest of the video features, such as the background, identity, and face pose, are preserved. The algorithm also works for TTS-generated speech segments essential for our case. 

\subsection{Data Generation Pipeline}
\label{section:datagen_pipeline}

\textbf{Data Collection Module: } Random videos are first collected from various online sources such as YouTube. These random videos introduce real-world variations a lipreader encounters in real life, such as variations in the head-pose of the speaker, speaker's distance from the camera (lipreader), speaker's complexion, and lip structure. We post-process these videos with a face-detection model to detect valid videos. Valid videos are single-identity front-facing talking head videos with no drastic pose changes. 
Speech utterances are generated using TTS models on vocabulary curated automatically from online sources.

\textbf{Audio-Video Alignment Module:} In our next step, we randomly select a pair of driving speech and a face video. To generate lip-synced videos using Wav2Lip, we match the video and speech utterance length by aligning them and then padding the speech utterance with silence. Naively aligning the speech utterance on the driving video can lead to residual lip movements, as shown in Fig.~\ref{fig:aud-vid-algn}, `Misaligned Video' row. Wav2Lip does not modify the lip movements in the driving video in the silent region. As a result, the output contains residual lip movements (indicated in the red box) from the original video. This can confuse and cause distress to the user learning to lipread. Our audio-video alignment module aligns the speech utterance on the video region with lip movements, as shown in Fig.~\ref{fig:aud-vid-algn}, `Aligned Video' row. This way, Wav2Lip naturally modifies the original mouth movements to correct speech-synced mouth movements while keeping the regions with no mouth movements untouched. 
We use lip-landmarks and the rate of change of the lip-landmarks between a predefined threshold of frames to detect mouth movements in the face videos. Once we have detected lip movements, we align the audio on the detected video region and add silences around the speech. 

\textbf{Data Generation: }The aligned speech utterance and the face video are passed through Wav2Lip. Wav2Lip modifies the lip movements in the original video and preserves the original head movements, background, and camera variations, thus allowing us to create realistic-looking synthetic videos in the wild. Overall pipeline is illustrated in Fig.~\ref{fig:main_pipeline}.

\section{Human Lipreading Training}
\label{sec:protocol-definition}

Lipreading is an involved process of recognizing speech from visual cues - the shape formed by the lips, teeth, and tongue. 
A lipreader may also rely on several other factors, such as the context of the conversation, familiarity with the speaker, vocabulary, and accent. Thus, taking inspiration from lipreading.org and readourlips.ca\def\thefootnote{$^8$}\footnote{\href{https://www.readourlips.ca/}{https://www.readourlips.ca/}}, we define three lipreading protocols for  conducting a user study to evaluate the viability of our platform - (1) lipreading on isolated words (WL), (2) lipreading sentences with context (SL), and (3) lipreading missing words in sentences (MWIS). These protocols rely on a lipreader's vocabulary and the role that semantic context plays in a person's ability to lipread.

\subsection{Lipreading on isolated Words (WL)}
\label{section:word_lipreading}

The ability to disambiguate different words through visual lip movements helps shape auditory perception and speech production. In \textbf{w}ord-\textbf{l}evel (WL) lipreading, the user is presented with a video of an isolated word being spoken by a talking head, along with multiple choices and one correct answer. When a video is played on the screen, the user must respond by selecting a single response from the provided multiple choices. Visually similar words (homophenes) are placed as options in the multiple choices to increase the difficulty of the task. The difficulty can be further increased by testing for difficult words - difficulty associated with the word to lipread, e.g., uncommon words are harder to lipread.
For the purpose of our study, we test the users only on the commonly known words. The multiple answer choices have been fixed to 5 options. An example of word-level lipreading is shown in Fig.~\ref{fig:quiz_examples} (a).

\subsection{Lipreading Sentences with Context (SL)}
\label{section:lipreading-sentence}

In \textbf{s}entence-\textbf{l}evel (SL) lipreading, the users are presented with (1) videos of talking heads speaking entire sentences and (2) the context of the sentences. The context acts as an additional cue to the mouthing of sentences and is meant to simulate practical conversations in a given context. According to~\cite{context-bib}, the context of the sentences can improve a person's lipreading skills. Context narrows the vocabulary and helps in the disambiguation of different words. We evaluate our users in two contexts - A) Introduction - `how are you?', `what is your name?', and B) Lipreading in a restaurant - `what would you like to order?'. Like WL lipreading, we provide the user with a fixed number of multiple choices and one correct answer. 
Apart from context, no other information is provided to the participants regarding the length or semantics of the sentence.
Fig.~\ref{fig:quiz_examples} (b) shows an example of sentence-level lipreading with context.

\subsection{Lipreading missing words in sentences (MWIS)}
\label{section:missing-word}
According to\def\thefootnote{$^9$}\footnote{\href{https://www.cdc.gov/ncbddd/hearingloss/parentsguide/building/speech-reading.html}{Speech Reading, Hearing Loss in Children $|$ CDC}}, an expert lipreader can discern only $40\%$ of a given sentence or $4-5$ words in a $12$ words long sentence. In this protocol, we try to emulate such an experience by \textbf{m}asking \textbf{w}ords \textbf{i}n the \textbf{s}entence (MWIS). 
The participants watch videos of sentences spoken by a talking head with a word in the sentence masked, as shown in Fig.~\ref{fig:quiz_examples} (c).
Unlike SL mentioned in Sec.~\ref{section:lipreading-sentence}, the users are not provided with any additional sentence context. 
Lip movements are an ambiguous source of information due to the presence of homophenes. This exercise thus aims to use the context of the sentence to disambiguate between multiple possibilities and guess the correct answer. For instance, given the masked sentence "a cat sits on the \{masked\}," a lipreader can disambiguate between homophenes `mat', `bat', and `pat' using the sentence context to select `mat'. 
The user must enter the input in text format for the masked word as shown in Fig.~\ref{fig:quiz_examples} (c). Minor spelling mistakes are accepted.

\section{User Study}
\label{section:main-user-study}
In this section, we explain the collective background of our participants, the types of videos used for the study, and the design of our testing platform. 

\subsection{Participants}
\label{section:participants}
We perform our study on $50$ participants with varying degrees of hearing loss with $32$ male and $18$ female participants. The average age of the participants in this study is $35$ years, ranging from $29$ years to $50$ years. 
Participants in this study reside in the Indian states of Maharashtra and Rajasthan. 
$29$ participants have a Master's degree, while the remaining $21$ have a Bachelor's degree. All the participants in the study report having sensorineural hearing loss\def\thefootnote{$^{10}$}\footnote{\href{https://www.healthline.com/health/sensorineural-hearing-loss}{What is Sensorineural Hearing Loss?}} and use hearing aids in their daily life along with lipreading and oral deaf speech as their primary mode of communication.

\subsection{Dataset}

\begingroup
\setlength{\tabcolsep}{8pt}
\renewcommand{\arraystretch}{1}
We scrape real videos from lipreading.org and generate our synthetic videos on them. Lipreading.org videos allow us to (i) make a direct comparison between the real lipreading training videos and our synthetically generated videos and (ii) provides the correct answer to the video; this provides the correct ground truth label for the real videos later used for quantitative analysis. 

Primarily, we aim to compare a user's performance on the synthetic videos generated using our proposed pipeline against the real videos on lipreading.org. We use the three protocols explained in Sec.~\ref{sec:protocol-definition} for this purpose. Our synthetic videos are divided into: (1) non-native \textbf{A}merican-accented \textbf{E}nglish (AE) videos and (2) native \textbf{I}ndian-accented \textbf{E}nglish (IE) videos. Our users are of Indian origin. 

\begin{table}[t]
  \centering
  \adjustbox{max width=0.75\linewidth}{
  \begin{tabular}{|l | c | c c |}
      \hline
      \multicolumn{1}{|c| }{} & \multicolumn{1}{c| }{Real} & \multicolumn{2}{ c|}{Synthetic}
      \\ \cline{2-4}
      Task & American & American & Indian
       \\ \hline
      WL &  80 & 800 & 800
      \\ 
      SL & 60 & 600 & 600
      \\ 
      MWIS & 70 & 700 & 700
      \\ \hline
      \textbf{Total} & 210 & 2100 & 2100
      \\ \hline
  \end{tabular}}
  \caption{\small No. of examples curated for each protocol in different English accents (American / Indian). }
  \label{tab:database_stats}
\end{table}
\endgroup

Our synthetic dataset is created using $10$ driving videos on $5$ speakers. We scrape $80$ labels from lipreading.org's single-word lipreading quiz for WL lipreading protocol. Using these, we generate $80 \times 10 = 800$ talking head videos - $10$ variations per word. For SL lipreading, we scrape $60$ questions from lipreading.org's sentence-level quiz across two contexts: introductions and lipreading in a restaurant.
We generate $60 \times 10 = 600$ talking head videos - $10$ variations for each sentence using these sentences. Lastly, we scrape $70$ sentences from lipreading.org's missing words in sentences task and generate $70 \times 10 = 700$ talking head videos for the MWIS protocol. We generate these videos once using American-accented TTS and the second time using Indian-accented TTS. As shown in Table~\ref{tab:database_stats}, we generate a total of $4200$ synthetic videos and collect $210$ real videos from lipreading.org across the protocols.

\begin{figure*}
\begin{center}
\includegraphics[width=\textwidth]{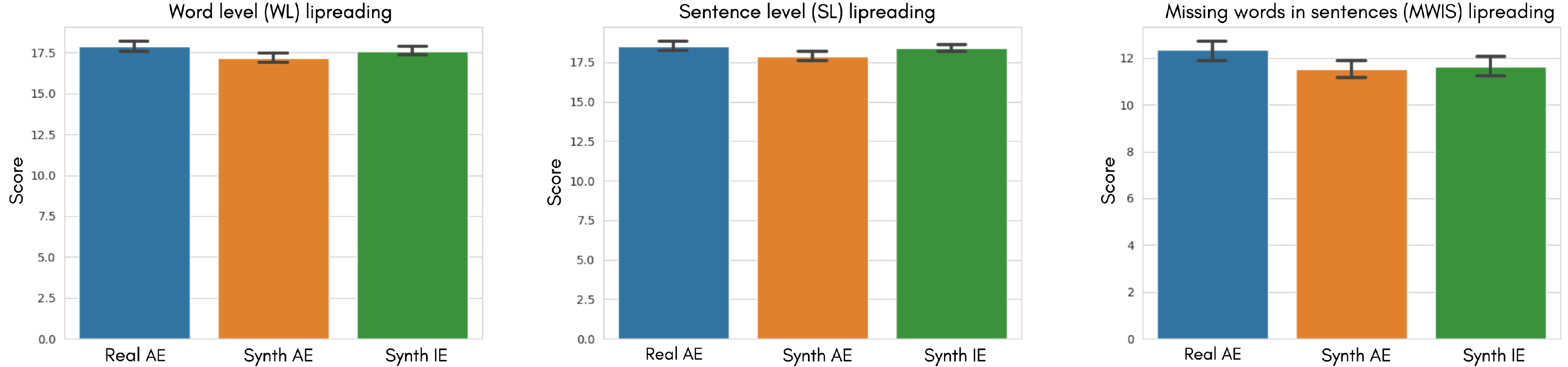}
\end{center}
   \caption{\small Mean user performance on the three lipreading protocols. Error bars are the standard errors of the mean.}
\label{fig:errorbar}
\end{figure*}

\begin{figure*}
\begin{center}
\includegraphics[width=\textwidth]{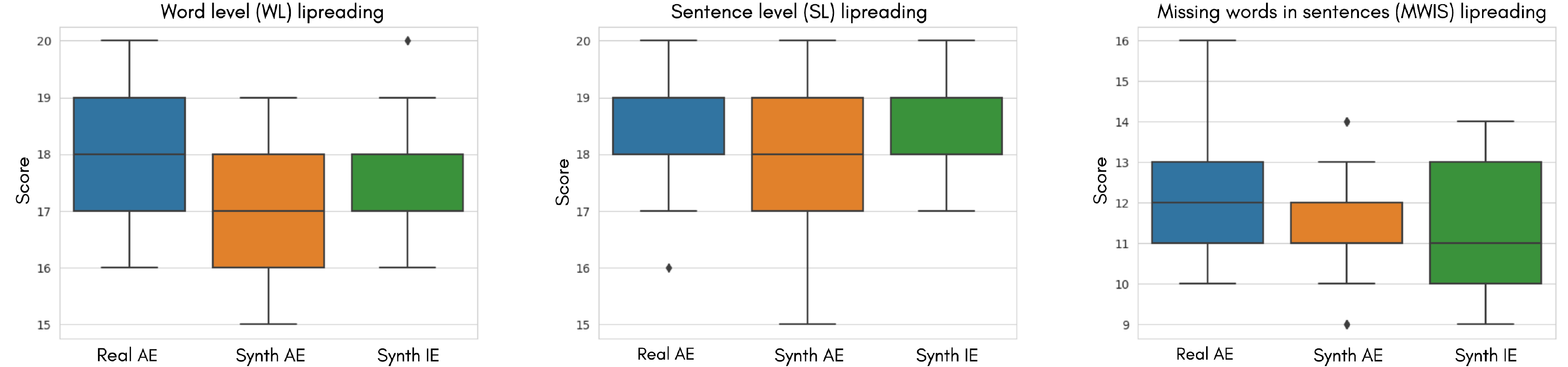}
\end{center}
   \caption{\small Box plots depicting the distribution of scores on the three lipreading protocols. Horizontal lines within the rectangles represent median scores. The top and bottom of the rectangles correspond to the first and third quartiles; the horizontal lines at the ends of the vertical “whiskers” represent the minimum and maximum scores, and the diamonds represent scores outside this range.}
\label{fig:boxplots}
\end{figure*}

\subsection{Test Design}
Our primary goal is to validate that the synthetic talking head videos generated using our pipeline can replace real videos in terms of visual quality and ease of discernment. 

Each participant participates in all $3$ protocols. For each protocol, the user takes $3$ quizzes corresponding to three datasets: (1) Real AE, (2) \textbf{Synth}etic \textbf{AE} (Synth AE), and (3) \textbf{Synth}etic \textbf{IE} (Synth IE). In total, a user attempts $9$ quizzes. Quizzes are delivered through a web-based platform that we developed. Our users report taking the quizzes from a plethora of personal devices like PCs, laptops, Android and iPhone mobile devices and tablets. The number of days taken to complete a test is left at the user's discretion to prevent the user from feeling fatigued, as lipreading is an involved process and can be mentally taxing. The longest time taken by any user to complete our test is four days.

The user is presented with $20$ questions/videos for each quiz. A word/sentence is first randomly sampled from the database for each question. One of the $10$ variations of the sampled word/sentence present in the database is then randomly chosen. The audio is removed from the videos before displaying to the users. 
We ensure that words/sentences are not repeated across the quizzes in a single protocol to prevent bias by familiarization. We also ensure that the difficulty of lipreading across all the datasets and protocols is kept consistent.
The user is rewarded $1$ point for each correct attempt, and the score is computed out of $20$. 
We expect the user to finish a single test in one sitting. For a fair comparison, we do not inform the user if they are being tested on real or synthetic data.


\section{Results and Discussion}
\label{section:results}

In this section, we conduct statistical analysis to verify \textbf{(T1)} If the lipreading performance of the users remains comparable across the real and synthetic videos generated using our pipeline. Through this, we will validate the viability of our proposed pipeline as an alternative to the existing online lipreading training platforms. \textbf{(T2)} If the users are more comfortable lipreading in their native accent/language than in a foreign accent/language. This would validate the need for bootstrapping lipreading training platforms in multiple languages/accents across the globe. 

Fig.~\ref{fig:errorbar} plots the standard errors of the mean. Fig.~\ref{fig:boxplots} presents the boxplot across the three lipreading protocols. 

\textbf{Synthetic videos as a replacement for real videos: }To validate \textbf{(T1)}, the difference in the user scores across the real and synthetic videos should be statistically insignificant. 
Since our conclusion depends on the evidence for a null hypothesis (no difference between the categories), just the absence of evidence is not enough to support the hypothesis. Therefore, we perform a Bayesian Equivalence Analysis using the Bayesian Estimation Supersedes the t-test (BEST)~\cite{best} to quantify the evidence in favor of our model. 
BEST estimates the difference in means between two distributions/groups and yields a probability distribution over the difference. Using this method, we compute (1) the mean credible value as the best guess of the actual difference between the two distributions and (2) the 95\% Highest Density Interval (HDI) as the range where the actual difference is with 95\% credibility. For the difference in the two distributions to be statistically significant, the difference in their mean scores should lie outside the 95\% HDI. 

We report the BEST statistics on Real AE and Synth AE studies for all three lipreading protocols in Table~\ref{tab:user-study-best}. We also report the t-statistic and p-value using the standard two-tailed t-test. From Table.~\ref{tab:user-study-best}, it is clear that the BEST statistic lies within the acceptable 95\% HDI for all three protocols indicating that the difference in the scores between the two groups is statistically insignificant. This suggests that our pipeline is a viable alternative to the existing manually curated talking-head videos. 

\textbf{Native vs Non-native accented lipreading: } To validate \textbf{(T2)}, the difference in the user scores between native and non-native accented English should be statistically significant. 
Since our participant pool is from India, we compare the user scores on Synth IE and Synth AE. We perform a two-sample Z test to validate the statistical significance since our sample size is large ($> 30$). To this end, we propose Null Hypothesis \textbf{H0}: the difference in the mean scores between Synth IE and Synth AE is statistically insignificant, and consequently, the Alternate Hypothesis \textbf{H1}: the difference in the mean scores between the Synth IE and Synth AE is statistically significant. We compute the z statistics and report the p-value for the 90\% confidence interval (significance value $\alpha$=0.1) in Table~\ref{tab:user-study-aevsie} for the three protocols. We observe that the Z test statistic lies outside the $90\%$ critical value accepted range for two tasks, WL and SL, indicating that the difference in their mean values is statistically significant in favor of IE, and we reject \textbf{H0} in favor of \textbf{H1} for these protocols. For MWIS protocol, the p-value is $> 0.1$, and the z statistic falls within the acceptable 90\% confidence interval, indicating that the difference in their mean scores is not statistically significant. Thus, we fail to reject \textbf{H0} in this case. 
The overall results support our claim that lipreading on native accents makes much difference in the performance of a lipreader, and they are more comfortable in lipreading native accents. Moreover, it reinforces the importance of our platform. 

Developing a lipreading training database for each new accent using real videos is a non-trivial, exhausting, and time-consuming task. Our platform could thus be quickly adopted to add any new language/accent as long as a TTS model for that language/accent is available.

\begingroup
\setlength{\tabcolsep}{7pt} 
\renewcommand{\arraystretch}{1.4} 
\begin{table}[t]
  \centering
  \adjustbox{max width=\linewidth}{
  \begin{tabular}{| l | c | c | c | c | c |}
    \hline
      & $\mathbf{95\%}$ \textbf{HDI} & \textbf{Mean} & \textbf{MGD} & \textbf{t-value} & \textbf{p-value}
      \\ \hline
      \textbf{WL}  & (-0.254, 1.63) & 0.701 & 0.706 & 1.676 & 0.103
      \\
      \textbf{SL} & (-0.226, 1.62) & 0.671 & 0.647 & 1.540 & 0.133
      \\ 
      \textbf{MWIS} & (-0.366, 1.98) & 0.793, & 0.824 & 1.517 & 0.139
      \\ \hline 
  \end{tabular}
  }
  \caption{\small We perform BEST statistical analysis and compute the 95\% HDI range of the difference in means of the real and synthetic distributions. Mean is the distribution of means. We also report the p-values and t-values from a standard t-test for comparison.}
  \label{tab:user-study-best}
\end{table}
\endgroup

\begingroup
\setlength{\tabcolsep}{7pt} 
\renewcommand{\arraystretch}{1.2} 
\begin{table}[t]
  \centering
  \adjustbox{max width=0.9\linewidth}{
  \begin{tabular}{| l | c | c | c |}
    \hline
      & \textbf{p-value} & \textbf{accepted range} & \textbf{z statistic}
      \\ \hline
      \textbf{WL}  & 0.0786 & (-1.645 : 1.645) & 1.758 
      \\
      \textbf{SL} & 0.0171 & (-1.645 : 1.645) & 2.384 
      \\ 
      \textbf{MWIS} & 0.705 & (-1.645 : 1.645) & 0.378
      \\ \hline 
  \end{tabular}
  }
  \caption{\small Two-sample z-test on synthetic Indian-accented English (IE) and American-accented English videos (AE). The significance level $\alpha$ is kept at $0.1$. The null-hypothesis is rejected if the z statistic falls outside the 90\% critical value accepted range. Consequently, the p-value is also less than the significance value $\alpha$ in that case. }
  \label{tab:user-study-aevsie}
\end{table}
\endgroup

\textbf{Discussion: }We note that the lipreaders score relatively higher for the SL protocol. The context of the sentence narrows the vocabulary space and helps disambiguate homophenes. 
MWIS is the most challenging protocol as it involves the user's needing to retrieve the correct word from their memory instead of classifying from the given choices. It also involves mapping the masked word from sentences to its corresponding mouthing in the videos. 
Thus, the users score relatively low on MWIS.

As a conclusion of the user study, we present evidence that synthetic videos can potentially replace real videos. We show that the drop in user performance across Real AE and Synth AE is statistically insignificant across all the protocols. We also show that users are more comfortable lipreading in a native accent through paired z-test, highlighting the dire need to bootstrap lipreading platforms in multiple languages/accents at scale.they

\section{CONCLUSION}

Lipreading is a widely adopted mode of communication for people with hearing loss. However, online resources for lipreading training are scarce and limited in many factors, such as vocabulary, speakers, and languages. Moreover, launching a new platform in a new language is costly, requiring months of manual effort to record training videos on hired actors.
In this work, we analyze the viability of using synthetically generated videos to replace real videos for lipreading training. We propose an end-to-end automated and cost-effective pipeline for generating lipreading videos and carefully design a set of protocols to evaluate the generated videos. We perform statistical analysis to validate that the difference in user performance on real and synthetic lipreading videos is statistically insignificant. We also show the advantage of lipreading in native accents, thus highlighting the dire need for lipreading training in many languages and accents. In this vein, we envision a MOOCs platform for training humans in lipreading to potentially impact millions of people with hearing loss across the globe.

{\small
\bibliographystyle{ieee_fullname}
\bibliography{ms}
}

\end{document}